\documentclass[letterpaper, 10 pt, conference]{ieeeconf}  

\IEEEoverridecommandlockouts                              

\usepackage{graphics} 
\usepackage{epsfig} 
\usepackage{subfigure}
\usepackage{mathptmx} 
\usepackage{amsmath} 
\usepackage{amssymb}  
\usepackage{makecell}
\usepackage{multirow}
\usepackage{float}
\usepackage[colorlinks=true,linkcolor=blue,citecolor=green,urlcolor=blue,]{hyperref}

\title{\LARGE \bf
PIEKF-VIWO: Visual-Inertial-Wheel Odometry using\\ Partial Invariant Extended Kalman Filter
}

\author{Tong Hua, Tao Li and Ling Pei$^*$
\thanks{*This work was supported in part by the National Nature Science Foundation of China (NSFC) under Grant 61873163, in part by the Shanghai Science and Technology Committee under Grant 20511103103, and in part by Equipment Preresearch Field Foundation under Grant 80913010303.}
\thanks{All authors are with Shanghai Key Laboratory of Navigation and Location Based Services, Shanghai Jiao Tong University. $^*$ Corresponding Author {\tt\small ling.pei@sjtu.edu.cn} }%
}

\begin{document}

\maketitle
\thispagestyle{empty}
\pagestyle{empty}

\begin{abstract}

Invariant Extended Kalman Filter (IEKF) has been successfully applied in Visual-inertial Odometry (VIO) as an advanced achievement of Kalman filter, showing great potential in sensor fusion. In this paper, we propose partial IEKF (PIEKF), which only incorporates rotation-velocity state into the Lie group structure and apply it for Visual-Inertial-Wheel Odometry (VIWO) to improve positioning accuracy and consistency. Specifically, we derive the rotation-velocity measurement model, which combines wheel measurements with kinematic constraints. The model circumvents the wheel odometer's 3D integration and covariance propagation, which is essential for filter consistency. And a plane constraint is also introduced to enhance the position accuracy. A dynamic outlier detection method is adopted, leveraging the velocity state output. Through the simulation and real-world test, we validate the effectiveness of our approach, which outperforms the standard Multi-State Constraint Kalman Filter (MSCKF) based VIWO in consistency and accuracy.
\newline
\par{Index Terms: } Invariant Extended Kalman Filter, sensor fusion, consistency
\end{abstract}

\section{INTRODUCTION}

Visual-Inertial Odometry (VIO) has recently been widely studied for its lightweight and high-precision advantage. Camera-IMU pose estimation is available in real commercial applications such as smart drones and mobile phones \cite{zhang2021pose}. However, when visual and inertial sensors are deployed on ground vehicles, degenerative scenes like stationary and constant linear acceleration motions and vehicle vibration noise often lead to significant challenges in pose estimation and system observability. Thus, it is necessary to aid VIO with additional sensors such as wheel encoders \cite{lee2020visual}.

Although the wheel measurements can improve the performance of VIO pose estimation, inconsistency may also negatively affect system state estimation. Experiments in \cite{lee2020visual} show that the online calibration of Visual-Inertial-Wheel Odometry (VIWO) will lead to high inconsistency with bad initial values. The analysis and improvement of VIO consistency have been mentioned in previous literature. Among them, Invariant Extended Kalman Filter (IEKF) maintains consistency theoretically and can be readily combined with other measurements such as the GPS. In contrast, other methods, such as the submap approach, are usually tailored \cite{barrau2015ekf}. Therefore, the introduction of invariance into VIWO is a better choice.

Meanwhile, the propagation matrix is highly related to the velocity and position state, which means IEKF heavily depends on the accuracy of noise parameters \cite{wang2021covariance}. Experiments in \cite{barrau2015ekf,hartley2020contact,wu2017invariant} are all conducted in the low-speed and small-scale environments. \cite{wang2021improved} has shown that unpredictable disturbances or substantial outliers may violate the basic assumptions of IEKF and lead to inaccurate approximation to the system model. Unfortunately, state estimation in outdoor scenarios usually suffers from dynamic scene points, unstructured environments with homogeneous and non-textured surfaces \cite{buczko2016distinguish}, which poses a great challenge to the IEKF model.

To address the challenge, a partial IEKF-based VIWO (PIEKF-VIWO) is developed, which integrates the wheel encoder measurements into invariant MSCKF. The model improves the consistency and robustness of VIWO compared with the conventional IEKF model in practice. Our main contributions are as follows:

\begin{itemize}
\item A partial IEKF (PIEKF) is derived and applied for MSCKF-based VIWO, whose Lie group structure includes only the rotation and velocity.
\item To fit the PIEKF framework, a rotation-velocity measurement model for the wheel odometer is proposed. 
\item To improve the robustness and accuracy, we introduce a PIEKF-based plane constraint model for the wheel odometer and a velocity-based dynamic outlier detection method for the camera.

\end{itemize}

\section{RELATED WORKS}

\subsection{Visual-Inertial-Wheel Odometry}

In recent years, a couple of pose estimation methods have incorporated wheel measurements into the visual-inertial navigation system \cite{wu2017vins,dang2018tightly,kang2019vins,quan2019tightly,jung2020monocular}. The scale is proved to be observable by incorporating wheel encoder data, and a plane motion constraint is introduced to achieve higher position accuracy \cite{wu2017vins}. Based on the observability analysis, \cite{liu2021bidirectional} proposes a bidirectional trajectory computation method that solves the unobservability before the first tuning. These works usually utilize relative pose constraints integrated by wheel measurements. Although the wheel pre-integration has high time efficiency, it loses velocity information and limits accuracy improvement. 

\cite{zhang2021pose,zhang2019vision} extend the wheel encoder application from planar surface to manifolds and achieve high-accuracy 6D pose estimation. Kinematic constraints based on the instantaneous centers of rotation (ICR) model \cite{martinez2005approximating} are introduced into the optimization model of visual-inertial localization in \cite{zuo2019visual}. Liu et al. present a tightly-coupled VIWO with an initialization module using wheel encoder readings and online extrinsic calibration \cite{liu2019visual}. Different strategies for motion tracking are provided in \cite{quan2019tightly} to exploit IMU and wheel encoder cues maximally. However, most of the work is based on the nonlinear optimization method, which is time-consuming and does not focus on consistency. Lee et al. propose a consistent VIWO along with both extrinsic and intrinsic calibration of the wheel encoder and analyze the observability \cite{lee2020visual}. However, this method requires complex covariance propagation to guarantee system consistency.

\subsection{Filter consistency}
For the extended Kalman filter framework, linearization errors can cause filter inconsistencies, leading to overly optimistic covariance estimates, which have been discussed in rich literature (e.g., \cite{bailey2006consistency, huang2007convergence, barrau2015ekf, huang2008analysis,li2013high, huang2010observability}). There are a couple of solutions for the inconsistency problem. One is modifying the EKF linearization point represented by the First-Estimates Jacobian (FEJ) method \cite{chen2022fej2,huang2009first} which guarantees that the null space of the observability matrix does not degenerate. OC-EKF \cite{hesch2012observability} explicitly enforces the system's observability constraints and maintains the null space. Another solution is robocentric odometry represented by ROVIO \cite{2015Robust} and R-VIO \cite{huai2018robocentric} which originates from robocentric mapping \cite{castellanos2004limits}. This method avoids the alignment with the global gravity vector and the observability mismatch for the world-centric VIO. 
In recent years, Bonnabel et al. propose IEKF based on a novel Lie group structure \cite{bonnabel2007left, bonnable2009invariant}. The consistency of this new scheme is demonstrated in \cite{barrau2015ekf} and further applied to MSCKF \cite{wu2017invariant, heo2018consistent} and UKF \cite{brossard2018invariant}. In particular, the invariant filter exhibits the potential for long-term inertial navigation when combined with some robust components \cite{brossard2019rins}.

\section{PRELIMINARY KNOWLEDGE}
Before describing our method, the coordinate frames and some notations involved are clarified in Table \ref{tab2}. The wheel odometer and body frames are equivalent by default in the following discussion.
\begin{table}[htp]
  \centering
  \caption{Glossary of notation.}
  \setlength{\tabcolsep}{0.8mm}{
  \renewcommand{\arraystretch}{1.9} 
  \begin{tabular*}{250pt}{cccc}
  \hline
  \hline    
  Symbol & Meaning & Symbol & Meaning \\
  \hline    
  $G$  & Global frame & $I$& IMU Frame \\

  \hline
  $O$ &Wheel odometer Frame & $C$& Camera Frame\\
  
  \hline
  ${}^B_A\mathbf{R}$ & \makecell[c]{Rotation from frame \\A to frame B} & ${}^B\mathbf{v}_A$&\makecell[c]{Frame A's velocity \\in frame B}\\

  \hline
  \makecell[c]{${}^B\mathbf{p}_A$}
  &\makecell[c]{Frame A's position \\in frame B}
  &$\mathbf{b}_g$&\makecell[c]{Gyroscope bias}\\ [1pt]
  
  \hline
  \makecell[c]{$\mathbf{b}_a$} & \makecell[c]{Acceleromter bias} & $\hat{\mathbf{x}} $ &\makecell[c]{Estimated value of $\mathbf{x}$} \\ [1pt]
  
  \hline
  $\Tilde{\mathbf{x}}$  & \makecell[c]{Error value of $\mathbf{x}$}
  & $\mathbf{e}_i$ & The $\it{i}$th column of $\mathbf{I}_3$ \\[1pt]
  
  \hline
  $\lfloor \mathbf{a} \rfloor$  & \makecell{Skew symmetric matrix \\of $\mathbf{a} \in \mathbb{R}^3$} & $exp(\cdot)$ & \makecell[c]{Mapping from $\mathbb{R}^3$ \\to the manifold SO(3)}\\
  
  \hline
  $log(\cdot)$  & \makecell[c]{Mapping from SO(3) \\to $\mathbb{R}^3$} & $\Vert \cdot \Vert$ & \makecell[l]{2-norm of the vector}\\
  \hline
  \hline
  \end{tabular*}}
  \label{tab2}
\end{table}

In the visual-inertial navigation system, the IMU state is given by:
\begin{equation}
\mathbf{X}_I = (
{ }_{I}^{G}\mathbf{R}, {}^G\mathbf{v}_I, { }^{G}\mathbf{p}_I, \mathbf{b}_g,\mathbf{b}_{a})\label{eq1}
\end{equation}
The IMU continuous kinematic model is given by:
\begin{equation}
\begin{aligned}
    {}^G_I\dot{\mathbf{R}} &= {}^G_I\mathbf{R}\lfloor {}^I\omega-\mathbf{b}_g-\mathbf{n}_g\rfloor \\
    {}^G\dot{\mathbf{v}}_I &= {}^G_I\mathbf{R}({}^I\mathbf{a}-\mathbf{b}_a-\mathbf{n}_a)+{}^G\mathbf{g}, {}^G\dot{\mathbf{p}}_I = {}^G\mathbf{v}_I\\
    \dot{\mathbf{b}}_g &= \mathbf{n}_{\omega g}, \dot{\mathbf{b}}_a = \mathbf{n}_{\omega a}
    \label{eq22}
\end{aligned}
\end{equation}
where ${}^I\omega$ and ${}^I\mathbf{a}$ are the angular velocity and linear acceleration velocity respectively. $\mathbf{n}_{g}$ and $\mathbf{n}_{a}$ are the Gaussian noise of the gyroscope and accelerometer measurements respectively, and $\mathbf{n}_{wg}$ and $\mathbf{n}_{wa}$ are the Gaussian noise of gyroscope and accelerometer measurement biases. 
Base on $SE_2(3)$ proposed in \cite{barrau2016invariant}, the IMU state on manifold can be constructed \cite{heo2018consistent}:
\begin{small}
\begin{equation}
    \chi_I = \begin{bmatrix}
    {}^G_I\mathbf{R} & {}^G\mathbf{v}_I & {}^G\mathbf{p}_I & & & & \\
            &  1 & & & & & \\
            & & 1 & & & & \\
     & &  & \mathbf{I}_3 & \mathbf{b}_g & &  \\
     & & & & 1 & & \\ 
     & & & & & \mathbf{I}_3 & \mathbf{b}_a \\
     & & & & & & 1
    \end{bmatrix} \label{eq3}
\end{equation}
\end{small}
We define the right invariant error $\xi_I \in \mathbb{R}^{15}$ by the following retraction:
\begin{equation}
    \chi_I = Exp(\xi_I)\hat{\chi_I} \label{eq4}
\end{equation}
where $Exp(\cdot)$ maps the vector $X \in \mathbb{R}^m $ into the manifold, and $\xi_I = \begin{bmatrix}
\xi_{\theta}^T & \xi_v^T & \xi_p^T & \xi_{b_g}^T & \xi_{b_a}^T
\end{bmatrix}^T$. And specifically the uncertainty representation of $\mathbf{X}_I$ is shown as follows:
\begin{small}
\begin{equation}
    \begin{aligned}
    \mathbf{X}_I = (&exp(\xi_\theta)\hat{\mathbf{R}}, exp(\xi_\theta)\hat{\mathbf{v}}+\mathbf J_l(\xi_\theta)\xi_v, exp(\xi_\theta)\hat{\mathbf{p}}+\mathbf J_l(\xi_\theta)\xi_p,\\
    &\mathbf{\hat{b}}_g + \xi_{b_g}, \mathbf{\hat{b}}_a + \xi_{b_a}) 
    \end{aligned}\label{eq31}
\end{equation}
\end{small}
where $\mathbf J_l(\cdot)$ is the left Jacobian for Lie Group. An invariant error state propagation equation is established.
\begin{equation}
    \dot{\xi}_I = \mathbf{F}\xi_I + \mathbf{G}\mathbf{n}_I \label{eq5}
\end{equation}
where $\mathbf{n}_I=\begin{bmatrix}\mathbf{n}_g^T & \mathbf{n}_{wg}^T & \mathbf{n}_a^T & \mathbf{n}_{wa}^T\end{bmatrix}^T$. The details of $\mathbf{F}$ and $\mathbf{G}$ can refer to \cite{wu2017invariant}. 

For MSCKF, the full state vector is augmented by the camera state:
\begin{equation}
    \mathbf{X} = (
    \mathbf{X}_I, \mathbf{X}_{C1}, ... , \mathbf{X}_{Cn}) \label{eq30}
\end{equation}
where $n$ is the number of camera states in the sliding window, and the camera state is represented by:
\begin{equation}
    \mathbf{X}_C = ({}^G_C \mathbf{R}, {}^G \mathbf{p}_C) \label{eq2}
\end{equation}
The application of invariant MSCKF on VIO can refer to \cite{wu2017invariant, heo2018consistent}.

\section{PIEKF}
\begin{flushleft}
\begin{figure}
    \includegraphics[scale = 0.45]{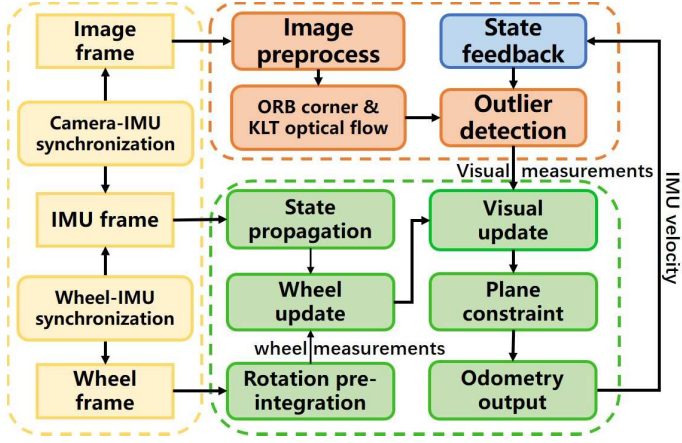}
    \caption{PIEKF-based VIWO system overview.}
    \label{fig_overview}
\end{figure}
\end{flushleft}

The main system components of PIEKF-based VIWO are shown in Fig. \ref{fig_overview}. In this section, we focus on the wheel measurement model and plane constraints with some details in the appendix. And the IMU-camera and IMU-wheel extrinsic calibration parameters are assumed to be perfectly known. 

\subsection{IMU kinematic model}
Unlike (\ref{eq31}), we only incorporate rotation and velocity into the Lie group structure, which means the filter has partial invariance. By decoupling the position with the rotation, we obtain a new uncertainty representation:
\begin{small}
\begin{equation}
    \begin{aligned}
    \mathbf{X}_I &= (exp(\xi_\theta)\hat{\mathbf{R}}, exp(\xi_\theta)\hat{\mathbf{v}}+\mathbf J_l(\xi_\theta)\xi_v, \hat{\mathbf{p}}+\xi_p,\mathbf{\hat{b}}_g + \xi_{b_g}, \mathbf{\hat{b}}_a + \xi_{b_a}) 
    \end{aligned}\label{eq33}
\end{equation}
\end{small}
And the corresponding propagation matrices $\mathbf{F}$ and $\mathbf{G}$ are given by:
\begin{small}
\begin{align}
\mathbf{F} &= \begin{bmatrix}
\mathbf{0}_{3\times 3} & \mathbf{0}_{3\times 3} & \mathbf{0}_{3\times 3} & -{}^G_I\hat{\mathbf{R}} & \mathbf{0}_{3\times 3}  \\
\lfloor {}^G\mathbf{g} \rfloor & \mathbf{0}_{3\times 3} & \mathbf{0}_{3\times 3} & -\lfloor {}^G\hat{\mathbf{v}}_{I} \rfloor {}^G_I\hat{\mathbf{R}} & -{}^G_I\hat{\mathbf{R}} \\
-\lfloor {}^G\mathbf{\hat{\mathbf{v}}}_{I} \rfloor & \mathbf{I}_{3} & \mathbf{0}_{3\times 3} & \mathbf{0}_{3\times 3} & \mathbf{0}_{3\times 3}  \\
\mathbf{0}_{3\times 3} & \mathbf{0}_{3\times 3} & \mathbf{0}_{3\times 3} & \mathbf{0}_{3\times 3} & \mathbf{0}_{3\times 3}  \\
\mathbf{0}_{3\times 3} & \mathbf{0}_{3\times 3} & \mathbf{0}_{3\times 3} & \mathbf{0}_{3\times 3} & \mathbf{0}_{3\times 3}
\end{bmatrix} \label{eq25}\\
\mathbf{G} &= \begin{bmatrix}
{}^G_I\hat{\mathbf{R}} & \mathbf{0}_{3\times 3} & \mathbf{0}_{3\times 3} & \mathbf{0}_{3\times 3} \\
\lfloor {}^G\hat{\mathbf{v}}_{I} \rfloor {}^G_I\hat{\mathbf{R}} & \mathbf{0}_{3\times 3} & {}^G_I\hat{\mathbf{R}} & \mathbf{0}_{3\times 3} \\
\mathbf{0}_{3\times 3} & \mathbf{0}_{3\times 3} & \mathbf{0}_{3\times 3} & \mathbf{0}_{3\times 3} \\
\mathbf{0}_{3\times 3} & \mathbf{I}_{3} & \mathbf{0}_{3\times 3} & \mathbf{0}_{3\times 3} \\
\mathbf{0}_{3\times 3} & \mathbf{0}_{3\times 3} & \mathbf{0}_{3\times 3} & \mathbf{I}_{3} 
\end{bmatrix}\label{eq26}
\end{align}
\end{small}
The difference from the standard IEKF model is that the propagation matrix in (\ref{eq25}) and (\ref{eq26}) does not contains the estimated position ${}^G\hat{\mathbf{p}_I}$ explicitly.

\subsection{Visual measurement model}
The camera state is represented as the manifold form $\chi_C$, which yields the state augmentation model:
\begin{align}
    \mathbf{P}_{aug} &= \begin{bmatrix}
    \mathbf{I}_{15+6n} \\
    \mathbf{J}
    \end{bmatrix}\mathbf{P}_{k+1|k}\begin{bmatrix}
    \mathbf{I}_{15+6n} \\
    \mathbf{J}
    \end{bmatrix}^T \label{eq23}
\end{align}
where the Jacobian $\mathbf{J}$ is written as:
\begin{equation}
    \mathbf{J} = \begin{bmatrix}
    \mathbf{I}_3 & \mathbf{0}_{3\times 3} & \mathbf{0}_{3\times 3} & \mathbf{0}_{3\times (6+6n)} \\
    -\lfloor {}^G_I\mathbf{R}{}^I\mathbf{p}_C \rfloor & \mathbf{0}_{3\times 3} & \mathbf{I}_3 & \mathbf{0}_{3\times (6+6n)}
    \end{bmatrix} \label{eq24}
\end{equation}
After initializing the positions of multiple observed features by the triangulation method, the normalized coordinate of the feature point ${}^G\mathbf{p}_f$ is obtained:
\begin{equation}
    \mathbf{z} = \pi({}^C\mathbf{p}_f) = \frac{1}{{}^C\mathbf{z}_f}\begin{bmatrix}
    {}^C\mathbf{x}_f \\ {}^C\mathbf{y}_f
    \end{bmatrix} = h(\mathbf{X},{}^G\mathbf{p}_f) \label{eq6}
\end{equation}
where $\pi$ is the projection function. The corresponding error vector of $\mathbf{X}$ is denoted as $\xi = \begin{bmatrix}\xi_I^T & \xi_{C1}^T & ... \xi_{Cn}^T\end{bmatrix}^T \in \mathbb{R}^{15+6n}$. After linearization, the visual measurement model can be obtained as below:
\begin{equation}
    \Tilde{\mathbf{z}} = \mathbf{H}_X\xi + \mathbf{H}_f{}^G\Tilde{\mathbf{p}}_f \label{eq7}
\end{equation}
The Jacobians $\mathbf{H}_x$ and $\mathbf{H}_f$ are given by:
\begin{align}
    \mathbf{H}_x &= \mathbf{J}_\pi\begin{bmatrix}
    \mathbf{0}_{2\times 15} & ... & \lfloor {}^C\mathbf{p}_{f} \rfloor{}^G_C\mathbf{R}^T & -{}^G_C\mathbf{R}^T & ...
    \end{bmatrix} \label{eq8} \\
    \mathbf{H}_f &= \mathbf{J}_\pi{}^G_C\mathbf{R}^T \label{eq9}
\end{align}
where $\mathbf{J}_\pi$ represents the Jacobian of $\pi$.

\subsection{Wheel measurement model}
\begin{figure}
    \centering
    \includegraphics[scale=0.4]{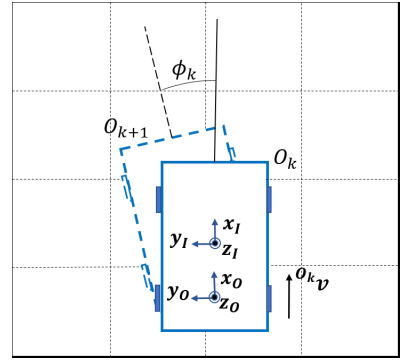}
    \caption{The vehicle and sensor measurement model.}
    \label{fig_wheel}
\end{figure}
Ground vehicles usually deploy two differential wheels with the 2D wheel odometer measurements introduced into the filter.
\begin{align}
    {}^{O_t} w &= \mathbf{e}_3^T{}^{O_t}\omega + n_\omega \label{eq10}\\
    {}^{O_t} v &= \mathbf{e}_1^T{}^{O_t}\mathbf{v} + n_v \label{eq11}
\end{align}
where ${}^{O_t}\omega$ and ${}^{O_t}\mathbf{v}$ are the angular velocity and linear velocity \cite{zhang2021pose} at time step $t$. As shown in Fig. \ref{fig_wheel}, the relative rotation measurement is obtained as follows:
\begin{equation}
    \phi_k = \int_{t_k}^{t_{k+1}}{}^{O_\tau}w d\tau
\end{equation}
Different from relative pose constraints in \cite{lee2020visual, wu2017vins, liu2019visual}, we adopt a rotation-velocity measurement model:
\begin{align}
    {}^{O_k}_{O_{k+1}}\theta &= \mathbf{e}_3^Tlog({}^O_I\mathbf{R}{}^{I_{k}}_G\mathbf{R}{}^G_{I_{k+1}}\mathbf{R}{}^I_O\mathbf{R}) \label{eq12}\\
    {}^{O_{k}}\mathbf{v} &= \lfloor{}^{I_{k}}\omega \rfloor {}^O_I\mathbf{R}{}^I\mathbf{p}_O + {}^{O_k}_G\mathbf{R}{}^G\mathbf{v}_{I_k} \label{eq13}
\end{align}
Using the right invariant error defined in (\ref{eq4}), the residual and measurement Jacobian are given as below:
\begin{align}
    \mathbf{r}_\theta &= \mathbf{e}_3^Tlog({}^{O_k}_{O_{k+1}}\hat{\mathbf{R}}^T{}^{O_k}_{O_{k+1}}\mathbf{R}) \label{eq14}\\
    \mathbf{H}_\theta &= \mathbf{e}_3^T\begin{bmatrix}
    {}^{O_{k+1}}_G\hat{\mathbf{R}} & \mathbf{0}_{3\times 12}& ... & -{}^{O_{k+1}}_G\hat{\mathbf{R}} & \mathbf{0}_{3\times 3}\end{bmatrix} \label{eq15}
\end{align}
where ${}^{O_k}_{O_{k+1}}\mathbf{R} = R_z(\phi_k)$ is the rotation measurement from the wheel odometer pre-integration. For velocity measurements, the non-holonomic constraints can be applied assuming that frame $O$ is the body frame, i.e., the velocity on the cross-track and vertical direction should be zero. Thus the 3D residual and corresponding Jacobian are obtained.
\begin{align}
    \mathbf{r}_v &= \begin{bmatrix}{}^{O_k}v & 0 & 0\end{bmatrix}^T - {}^O_I\mathbf{R}\lfloor {}^{I_k}\omega \rfloor {}^I\mathbf{p}_O - {}^{O_k}_G\hat{\mathbf{R}}{}^G\hat{\mathbf{v}}_{I_k} \label{eq16}\\
    \mathbf{H}_v &= \mathbf{e}_3^T\begin{bmatrix}
    \mathbf{0}_{3\times 3} & {}^{O_k}_G\hat{\mathbf{R}} & \mathbf{0}_{3\times 3} & {}^O_I\mathbf{R}\lfloor {}^I\mathbf{p}_{O} \rfloor & \mathbf{0}_{3\times (3+6n)}
    \end{bmatrix} \label{eq17}
\end{align}
The Jacobian in (\ref{eq17}) is irrelevant to rotation error, which is different from the conventional measurement model \cite{shin2002accuracy}. Additionally, velocity measurement update circumvents the 3D integration of wheel measurements and the covariance propagation compared with the relative displacement measurements in \cite{lee2020visual}.

\subsection{Plane constraint}
Assuming that the vehicle is running on the plane $\pi$ which is the initial x-y plane of frame $G$ \cite{wu2017vins}, the rotational and translational constraint is obtained:
\begin{small}
\begin{align}
    \mathbf{z}_{rot} &= \Lambda{}^{\pi}_G\mathbf{R}{}^G_{O_k}\mathbf{R}\mathbf{e}_3 \label{eq18} \\
    \mathbf{z}_{tran} &= \mathbf{e}_3^T{}^{\pi}_G\mathbf{R}{}^G\mathbf{p}_{O_k} \label{eq19}
\end{align}
\end{small}
where $\Lambda = \begin{bmatrix} \mathbf{e}_1 & \mathbf{e}_2\end{bmatrix}^T$.The corresponding Jacobians are given as below:
\begin{small}
\begin{align}
    \mathbf{H}_{rot} &= \Lambda \begin{bmatrix}
    -{}^{\pi}_G\mathbf{R}{}\lfloor {}^G_{O_k}\mathbf{R}\mathbf{e}_{3} \rfloor & \mathbf{0}_{3\times (12+6n)}
    \end{bmatrix}
    \label{eq20} \\
    \mathbf{H}_{tran} &= \mathbf{e}_3^T
    \begin{bmatrix}
    -{}^{\pi}_G\mathbf{R}\lfloor {}^G_{I_k}\mathbf{R}{}^I\mathbf{p}_O \rfloor & \mathbf{0}_{3 \times 3} & {}^{\pi}_G\mathbf{R} & \mathbf{0}_{3 \times (6+6n)}
    \end{bmatrix}
    \label{eq21}
\end{align}
\end{small}
The plane constraint update is performed after removing the redundant camera states.

\section{VISUAL OUTLIER DETECTION}
In the visual frontend, we preprocess the image using the contrast limited adaptive histogram equalization \cite{zuiderveld1994contrast} (CLAHE) method to enhance the contrast of images. And the ORB corners are tracked based on KLT optical flow algorithm \cite{lucas1981iterative}. To detect outliers from tracked features, epipolar constraint is a common criterion used in many VIOs \cite{sun2018robust, qin2018vins}. However, this constraint may fail if the translations of the dynamic outliers and the vehicle are close to the same plane. Before describing our outlier detection method, we have three assumptions when the vehicle is approximately running in a straight line:
\begin{itemize}
    \item The vehicle has a small rotation in two consecutive camera frames, i.e., the angular velocity ${}^I\omega \approx \mathbf{0}$.
    \item The extrinsic sensor parameters are reliable.
    \item The outlier has a constant velocity ${}^G\mathbf{v}_f$ between the two frames.
\end{itemize}

We start from the feature's 3D position in the camera frame:
\begin{equation}
    {}^C\mathbf{p}_f = \begin{bmatrix} X_f & Y_f & Z_f \end{bmatrix}^T = {}^C_G \mathbf{R}({}^G\mathbf{p}_f - {}^G\mathbf{p}_C) \label{eq34}
\end{equation}
Thus the coordinate in the normalized image plane is $\mathbf{x}_f = \begin{bmatrix} \frac{X}{Z} \frac{Y}{Z} 1\end{bmatrix}^T$. The velocity of feature on the normalized image plane is approximated as:
\begin{equation}
    \mathbf{v}_{n} \approx \frac{\Lambda({}^Cv_{f,z}\mathbf{p}_f - {}^C\mathbf{v}_f)}{Z_f} \label{eq35}
\end{equation}
where ${}^C\mathbf{v}_f = {}^C_G \mathbf{R}({}^G\mathbf{v}_f - {}^G\mathbf{v}_C)$, ${}^Cv_{f,z}$ is the z-axis component of ${}^C\mathbf{v}_{f}$. $Z_f$ can be estimated easily by the triangulation method, and supposing each feature is static, i.e., ${}^C\mathbf{v}_{f,est} = -{}^C_G\mathbf{v}_C$, the estimated velocity $\mathbf{v}_{n,est}$ on the normalized plane can be obtained. The measured velocity can be approximated by:
\begin{equation}
    \mathbf{v}_{m} = \frac{\mathbf{x}_f^{'}-\mathbf{x}_f}{\Delta t} \label{eq36}
\end{equation}
where $\mathbf{x}_f$ and $\mathbf{x}_f^{'}$ are the normalized coordinates in two consecutive frames respectively, and $\Delta t$ is the time step. The velocity error is evaluated as the criterion.
\begin{equation}
    \Delta v = ||\mathbf{v}_{m} - \mathbf{v}_{n,est}|| \label{eq37}
\end{equation}
\begin{table}[t]
    \caption{Simulation noise configurations.}
    \renewcommand{\arraystretch}{1.5} 
    \label{tab1}
    \centering
    \begin{tabular}{ll}
        \hline
        \hline
        \makecell[l]{Parameter} & \makecell[l]{Value} \\
        \hline

        \makecell[l]{Gyroscope noise}  & \makecell[l]{0.01$rad/s/\sqrt{Hz}$} \\
        \hline
        \makecell[l]{Acceleration noise} & \makecell[l]{0.01$m/s^2/\sqrt{Hz}$}\\
        \hline
        \makecell[l]{Gyroscope random walk} & \makecell[l]{0.0001$rad/s^2/\sqrt{Hz}$}\\
        \hline
        \makecell[l]{Acceleration random walk} & \makecell[l]{0.0001$m/s^3/\sqrt{Hz}$}\\
        \hline
        \makecell[l]{Wheel velocity noise} & \makecell[l]{0.1$m/s$}\\
        \hline
        \makecell[l]{Wheel angular velocity noise} & \makecell[l]{0.001$rad/s$}\\
        \hline
        \makecell[l]{Feature noise} & \makecell[l]{1 pixel} \\
        \hline
        \hline
    \end{tabular}
\end{table}
\begin{figure}[b]
  \centering
  \includegraphics[scale=0.55]{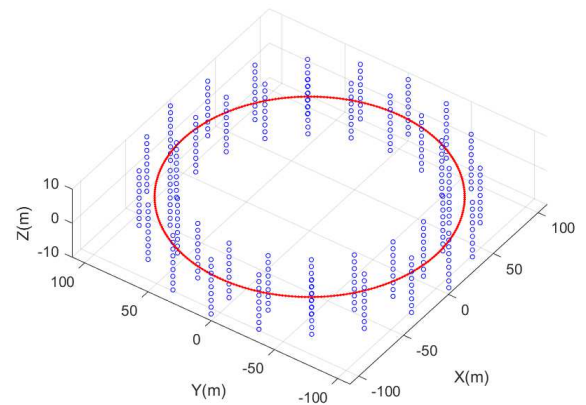}
  \caption{The error-free landmarks (blue) and trajectory (red).}
  \label{fig_gt}
\end{figure}
\begin{figure*}[t]
\subfigure[]{
    \centering
    \includegraphics[scale=0.5]{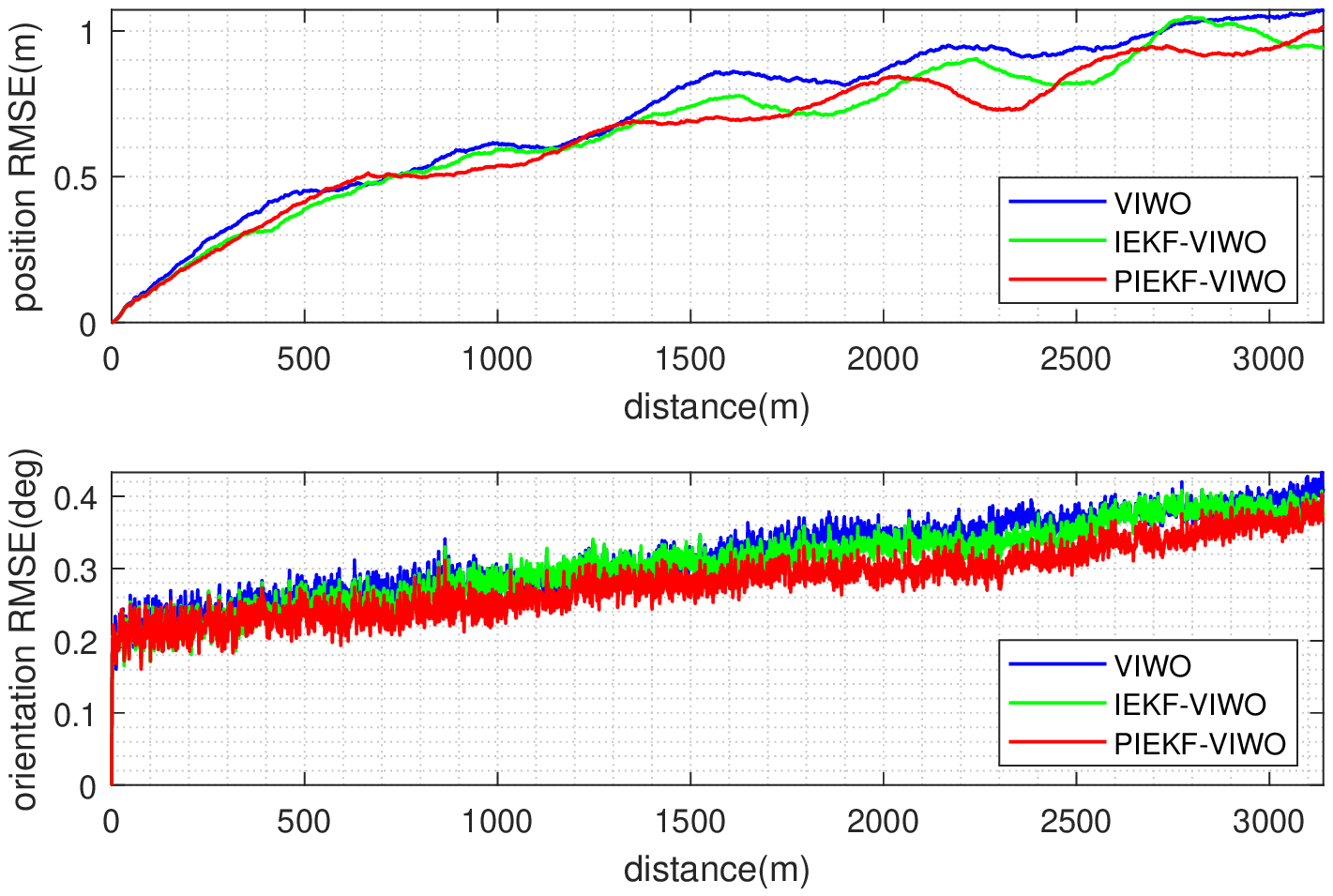}
}
\subfigure[]{
    \centering
    \includegraphics[scale=0.5]{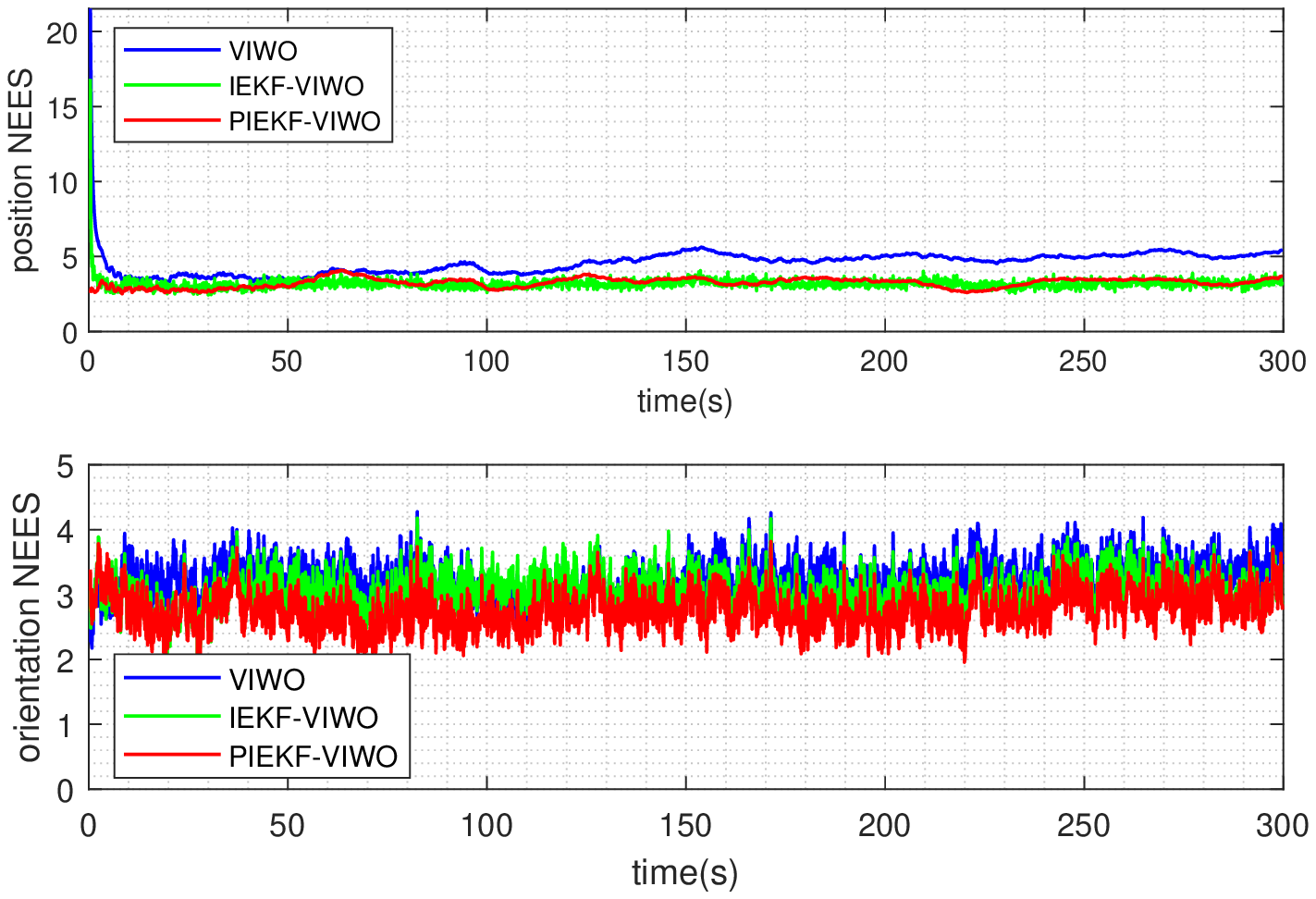}
}
\caption{Monte Carlo simulation results. (a): RMSE; (b): ANEES}
\label{fig_nees_rmse}
\end{figure*}
For the static feature, $\Delta v$ is close to zero, and for the dynamic outlier, the value depends on ${}^C\mathbf{v}_f$ and $Z_f$. Therefore we distinguish the outliers by setting the square mean value of all the errors as an adaptive threshold. It is noted that our method does not use the epipolar constraint, so it still works in the scene under the special motion.
\section{EVALUATION}
Both simulations and real-world tests are conducted to validate our proposed method. In the simulation test, we confirm the consistency and position accuracy of our proposed filter and evaluate the effectiveness of outlier detection. In the real-world test, we compare the position accuracy with other filter-based algorithms on Kaist urban dataset \cite{jeong2019complex}, and evaluate the plane constraints and time efficiency. Apart from PIEKF-VIWO, we have implemented the MSCKF-based VIWO (denoted as VIWO) and IEKF-based VIWO (denoted as IEKF-VIWO), both of which are the same as PIEKF-VIWO except the filtering framework. Our experiments are conducted on a laptop with Intel(R) Core(TM) i7-10710U CPU@1.10Ghz and 16G RAM.
\subsection{Simulation}
\begin{figure}[b]
\subfigure[]{
    \centering
    \includegraphics[scale = 0.27]{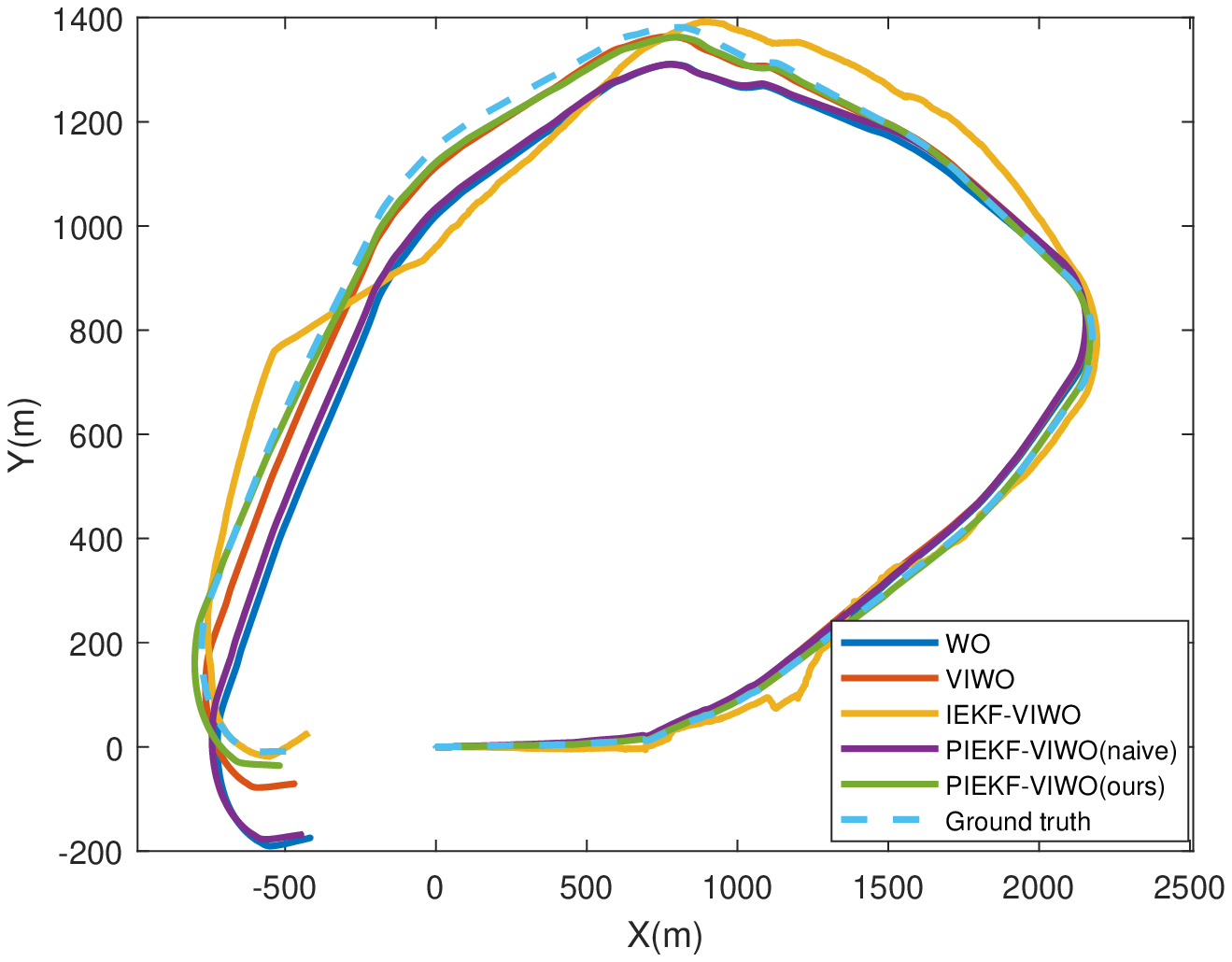}
}
\subfigure[]{
    \centering
    \includegraphics[scale = 0.27]{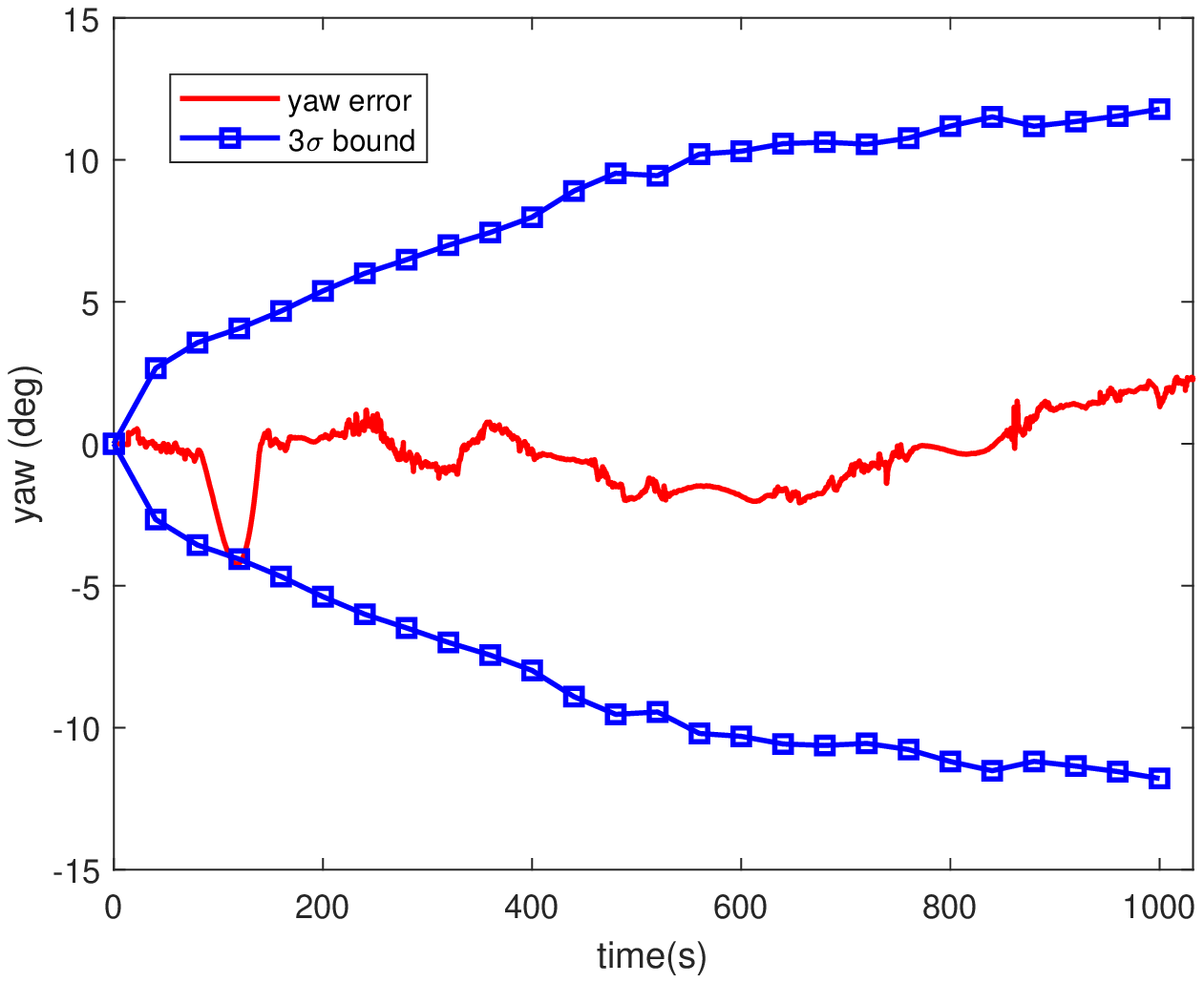}
}
\caption{Top view of estimated trajectories of algorithms (a) and yaw error with 3$\sigma$ bound (b) for PIEKF-VIWO in Kaist $urban32$.}
\label{fig_traj}
\end{figure}
\begin{figure}[b]
    \centering
    \includegraphics[scale = 0.45]{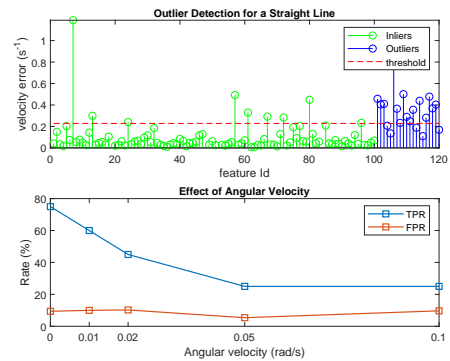}
    \caption{Outlier detection for simulated features. Features 1-100 are static features, and 101-120 are dynamic features. True Positive Rate (TPR) and False Positive Rate (FPR) are compared for different angular velocities.}
    \label{fig_dyna}
\end{figure}
\subsubsection{Consistency and accuracy}
We assume that a car equipped with 100Hz IMU, 100Hz wheel odometer and 10Hz camera is moving along a circle on a plane with 360 landmarks scattered around the true trajectory shown in Fig. \ref{fig_gt}. We conduct the Monte Carlo simulation for 50 runs where sensor measurements are generated with random noise following the noise distribution in Table \ref{tab1}. We compare VIWO, IEKF-VIWO, and our proposed PIEKF-VIWO by the accuracy indicator Root Mean Squared Error (RMSE) and the consistency indicator Averaged Normalized Estimation Error Squared (ANEES). The initial pose covariance for three algorithms is set to zero, and the results are shown in Fig. \ref{fig_nees_rmse}.
\begin{table*}[t]
    \caption{Orientation(deg)/Position(m) RMSE on the Kaist urban dataset.}
    \label{tab3}
    \centering
    \renewcommand\arraystretch{1.5}
    \begin{tabular}{ccccccc}
        \hline
        \multirow{2}*{Sequence(urban-)} & \multicolumn{6}{c}{Algorithm} \\
        \Xcline{2-7}{0.4pt}  & \makecell[c]{WO} & \makecell[c]{VIO} & \makecell[c]{VIWO}  & \makecell[c]{IEKF-VIWO}  & \makecell[c]{PIEKF-VIWO(naive)} & \makecell[c]{PIEKF-VIWO(ours)} \\
        \hline
        
        \makecell[c]{29(3.2km)} & \makecell[c]{4.13/13.72} & \makecell[c]{4.14/119.07}  & \makecell[c]{\underline{0.81}/\underline{2.41}}  & \makecell[c]{1.27/9.16}  & \makecell[c]{1.44/7.59}  & \makecell[c]{$\mathbf{0.68}$/$\mathbf{1.48}$} \\

        \makecell[c]{30(4.7km)} & \makecell[c]{16.64/145.47} & \makecell[c]{63.25/409.59}  & \makecell[c]{\underline{2.11}/$\mathbf{7.57}$}  & \makecell[c]{16.48/167.30}  & \makecell[c]{4.25/20.26}  & \makecell[c]{$\mathbf{1.76}$/\underline{8.39}} \\
        
        \makecell[c]{31(10.7km)} & \makecell[c]{14.47/363.05} & \makecell[c]{22.54/811.06}  & \makecell[c]{\underline{8.25}/\underline{177.97}}  & \makecell[c]{16.01/494.75}  & \makecell[c]{24.45/477.40}  & \makecell[c]{$\mathbf{5.21}$/$\mathbf{91.77}$} \\
        
        \makecell[c]{32(6.4km)} & \makecell[c]{4.00/104.84} & \makecell[c]{$\times$/$\times$} & \makecell[c]{\underline{2.14}/\underline{44.17}}  & \makecell[c]{12.15/74.88}  & \makecell[c]{3.50/92.66}  & \makecell[c]{$\mathbf{1.36}$/$\mathbf{19.03}$} \\
        
        \makecell[c]{33(7.3km)} & \makecell[c]{10.96/67.34} & \makecell[c]{$\times$/$\times$}  & \makecell[c]{\underline{4.00}/\underline{29.52}}  & \makecell[c]{8.54/79.62}  & \makecell[c]{7.11/87.88}  & \makecell[c]{$\mathbf{2.94}$/$\mathbf{27.36}$} \\
        
        \makecell[c]{34(6.5km)} & \makecell[c]{4.97/85.19} & \makecell[c]{$\times$/$\times$}  & \makecell[c]{$\mathbf{3.98}$/\underline{52.54}}  & \makecell[c]{7.57/80.57}  & \makecell[c]{5.75/58.58}  & \makecell[c]{\underline{4.72}/$\mathbf{51.97}$} \\
        
        \makecell[c]{35(3.2km)} & \makecell[c]{3.55/\underline{4.48}} & \makecell[c]{13.74/253.85}  & \makecell[c]{$\mathbf{1.19}/\mathbf{1.65}$}  & \makecell[c]{4.91/59.07}  & \makecell[c]{5.02/33.40}  & \makecell[c]{\underline{2.30}/9.07} \\

        \hline
        \multicolumn{7}{l}{$^{\mathrm{1}}$ $\times$ means the failure. Bold and underline indicate best and second best in each sequence, respectively.} 
    \end{tabular}
\end{table*}
Compared with VIWO and IEKF-VIWO, our proposed method has a slight gain in pose accuracy, whose RMSE is 0.648 m and 0.283 deg, respectively. The ideal NEES for orientation and position is three, and it is shown in Fig. \ref{fig_nees_rmse}(b) that PIEKF-VIWO performs similarly to IEKF-VIWO in consistency, better than the conventional VIWO especially in position NEES. 
\subsubsection{Outlier detection evaluation}
We simulate a scenario with 100 static features and 20 dynamic features around the trajectory. The vehicle moves at 54 km/h and each dynamic feature moves at a velocity from a zero mean Gaussian distribution with a standard deviation $\sigma_v = 20$ m/s along each axis per frame. We sample the initial angular velocity from 0 to 0.1 rad/s. The outlier detection result for the first frame is depicted in Fig. \ref{fig_dyna}. In the straght line scenario (Fig. \ref{fig_dyna}(a)), there are a total of 116 features tracked, among which 15 dynamic features are detected with a high true positive rate (75\%). Fig. \ref{fig_dyna}(b) demonstrates the detection effect is acceptable for this lightweight method when the angular velocity is small enough.
\begin{figure}
    \centering
    \includegraphics[scale = 0.5]{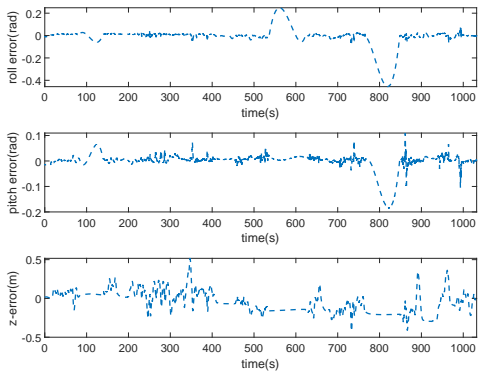}
    \caption{Roll, pitch and z-axis position error for PIEKF-VIWO on Kaist $urban32$.}
    \label{fig_z-error}
\end{figure}
\subsection{Real-world test}
To evaluate the performance of our algorithm in the real scenario, KAIST urban 29-35 are chosen for the test, which is collected with the 100Hz IMU, 100Hz wheel encoder, and a 10Hz left camera. The pose computed from the graph SLAM is used as the ground truth. We select the image frame when the vehicle first stops as the starting frame as MSCKF requires a static initialization except for $urban35$. Since $urban35$ does not contain a stationary period, we have to set the initial vehicle velocity to the velocity measured by the wheel encoder. We choose Wheel Odometry (WO), monocular VIO \cite{mourikis2007multi} and PIEKF-VIWO without outlier detection and plane constraints (PIEKF-VIWO(naive)) as benchmarks. 
\subsubsection{Accuracy and consistency}
Table \ref{tab3} lists the RMSE for different algorithms, and the estimated trajectories of $urban32$ are depicted in Fig. \ref{fig_traj}(a) as an example. It is noted that, unlike the simulation, IEKF-VIWO fails to achieve a good performance, especially in the dataset sequence with long-distance traveling and high speed, such as $urban31$. With the help of the PIEKF framework and additional constraints, PIEKF-VIWO is less vulnerable to noise and performs much better than IEKF and PIEKF-VIWO(naive). It is noted that PIEKF-VIWO is badly-behaved in $urban35$ because of the dynamic initialization. But for most cases, PIEKF-VIWO shows better results regarding both position and orientation accuracy than the conventional VIWO. The estimated yaw result in $urban32$ is also plot in Fig. \ref{fig_traj}(b), where the error remains in the 3$\sigma$ boundary, demonstrating the consistency of our algorithm.
\subsubsection{Plane constraint evaluation}
We demonstrate the effect of plane constraints in the partial IEKF algorithm. Real trajectories may have bumpy sections, so we relax the measurement error covariance to 0.01. The estimation error tested on the $urban32$ is depicted in Fig. \ref{fig_z-error}. The estimated roll, pitch, and position along the z-axis are maintained at around zero except in some sections where the constraints are not satisfied.
\subsubsection{Runtime comparison}
Since the algorithm performs wheel measurement updates at a high frequency, we compare the efficiency of our algorithms against VIO by computing the averaged backend processing time per frame for all the tested sequences, which is 2.46 (VIO), 5.84 (IEKF-VIWO) and 5.89 (PIEKF-VIWO) (unit: ms) respectively. IEKF-VIWO and PIEKF-VIWO have approximately the same efficiency in sequence as expected. Though our algorithms consume about twice as much time as VIO, their time consumption remains within 6ms, ensuring real-time in the test.

\section{CONCLUSIONS}

This paper proposes a PIEKF-based state estimation algorithm fusing camera, IMU, and wheel odometer. We compare our algorithm with conventional VIWO and demonstrate its better consistency and accuracy through the simulation and real-world experiments. In the future, our work will enhance robustness by introducing online calibration and improving the visual frontend.

\bibliographystyle{IEEEtran}
\bibliography{ref.bib}

\end{document}